# Augmented Lagrangian-Based Safe Reinforcement Learning Approach for Distribution System Volt/VAR Control

Guibin Chen*

*Abstract*—This paper proposes a data-driven solution for Volt-VAR control problem in active distribution system. As distribution system models are always inaccurate and incomplete, it is quite difficult to solve the problem. To handle with this dilemma, this paper formulates the Volt-VAR control problem as a constrained Markov decision process (CMDP). By synergistically combining the augmented Lagrangian method and soft actor critic algorithm, a novel safe off-policy reinforcement learning (RL) approach is proposed in this paper to solve the CMDP. The actor network is updated in a policy gradient manner with the Lagrangian value function. A double-critics network is adopted to synchronously estimate the action-value function to avoid overestimation bias. The proposed algorithm does not require strong convexity guarantee of examined problems and is sample efficient. A two-stage strategy is adopted for offline training and online execution, so the accurate distribution system model is no longer needed. To achieve scalability, a centralized training distributed execution strategy is adopted for a multi-agent framework, which enables a decentralized Volt-VAR control for large-scale distribution system. Comprehensive numerical experiments with real-world electricity data demonstrate that our proposed algorithm can achieve high solution optimality and constraints compliance.

## I. Introduction

With the integration of large-scale distributed energy resources (DERs), modern distribution networks face significant operational challenges, such as voltage violations and excessive active power losses. Reactive voltage control, as a critical component of distribution management systems (DMS), is employed to optimize the control actions of all voltage regulation and reactive power control devices (e.g., voltage regulators, load tap changers, and switchable capacitor banks) to minimize system losses and operational costs, while adhering to operational constraints such as voltage limits and line loading restrictions. Effectively utilizing the integrated large-scale DERs to achieve voltage regulation and reduce line losses is a pressing issue in the operation and control of modern power grids. In particular, with the increasing penetration of inverter-based energy resources (IB-ERs), which can provide rapid reactive power support, leveraging IB-ERs for reactive voltage control has attracted widespread attention.

Conventional reactive voltage control methods are primarily optimization-based approaches that rely on physical models of the power grid. Most existing reactive voltage control algorithms, whether centralized or distributed, depend on real-time measurement data and accurate physical system models of the grid. Common centralized reactive voltage control algorithms include well-known techniques such as the conic relaxation method [2], interior point method [3], mixed-integer linear programming [4], and evolutionary algorithms [5]. Distributed reactive voltage control algorithms, on the other hand, allow for communication and information sharing between neighboring control centers to collaboratively achieve global reactive voltage control objectives. Distributed algorithms in this domain include the alternating direction method of multipliers (ADMM) [6], dual decomposition methods [7], and accelerated ADMM [8], among others. To address the uncertainties introduced by the integration of distributed energy resources, methods such as model predictive control (MPC) [9] and scenario-based uncertainty optimization [10] have also been employed to tackle the reactive voltage control challenge.

However, the network parameters in the aforementioned methods are based on theoretical values rather than actual operational data from the grid. This model mismatch reduces the effectiveness of the solutions, hindering the practical application of model-based reactive voltage control algorithms in real-world scenarios. To eliminate the dependence on complete and accurate grid topology and parameter information, recent research has shifted focus towards using reinforcement learning (RL) algorithms to address reactive voltage control problems. RL-based reactive voltage control algorithms are model-free optimization methods that can learn optimal control strategies solely through measurement data and continuous exploration in the action space [11]. For instance, reference [12] proposed using the Q-learning algorithm to solve the optimal reactive power dispatch problem, while reference [13] introduced the use of radial basis functions to approximate the Q-function, enabling optimal tap position selection for voltage regulation devices. Moreover, reference [14] applied deep Q-learning to optimize capacitor switching in power systems for voltage regulation. However, deep Q-learning is limited to discrete action spaces and cannot effectively address control problems in continuous action spaces.

The RL-based reactive voltage control strategies achieve model-free optimization but face the following limitations: (1) In RL-based reactive voltage control methods, online training of the agent poses significant risks to system operations; (2) The action-value function, which is typically approximated using neural networks, suffers from inevitable estimation errors. This issue is particularly exacerbated when temporal difference learning is used for network training, leading to the accumulation of overestimation bias and subsequently hindering the optimization of the policy network; (3) Handling voltage violation constraints by modifying the reward function introduces penalization terms whose coefficients must be manually tuned based on training performance, potentially resulting in overly conservative policies or convergence to infeasible solutions.

To address these challenges, this paper models the reactive voltage control problem in distribution networks as a constrained Markov decision process (CMDP) and proposes a novel safe off-policy deep reinforcement learning (RL) algorithm, the Augmented Lagrangian-Soft Actor-Critic (AL-SAC). This approach directly regularizes the optimization problem by incorporating physical constraints. By leveraging the augmented Lagrangian method, we transform the CMDP problem with continuous action spaces into an unconstrained saddle-point optimization problem, allowing for automatic identification of optimal primal and dual variables. Compared to the commonly used primal-dual methods in safe deep reinforcement learning, the augmented Lagrangian method does not require strong convexity conditions of the problem and improves tractability while ensuring convergence guarantees. Additionally, the proposed algorithm employs maximum entropy regularization, which makes it more robust to hyperparameter settings and enables it to balance the exploration-exploitation trade-off, avoiding local optima. The algorithm utilizes a Double-Critic network architecture to estimate the expected future rewards for given actions, mitigating overestimation bias and local optimum issues often encountered in deep reinforcement learning training. The contributions of this paper are summarized as follows:

1) Instead of modifying the reward function of the Markov decision process (MDP) by adding penalty terms, this paper formulates the reactive voltage optimization problem as a constrained Markov decision process (CMDP) to explicitly model the voltage violation constraints in distribution networks. By synergistically combining the augmented Lagrangian method with the Soft Actor-Critic (SAC) algorithm, the proposed AL-SAC algorithm demonstrates superior performance in terms of both optimality and constraint satisfaction.

2) A two-stage safe deep reinforcement learning approach is proposed to enhance the safety and efficiency of online operations through offline pretraining of the model. The proposed augmented Lagrangian-based safe reinforcement learning method for reactive voltage control operates in a model-free manner during the online phase, while in the offline phase, it leverages an approximate model and historical data for pretraining. This design improves both the safety and efficiency of online learning, making the method more practical and effective for real-world applications.

3) This paper further extends the proposed augmented Lagrangian-based safe reinforcement learning method for reactive voltage control from a single-agent framework to a multi-agent framework, demonstrating the feasibility of deploying the proposed data-driven approach in larger-scale networks.

The remainder of this paper is as follows: Section Ⅱ provides a brief introduction to the relevant concepts and definitions, followed by the formulation of the reactive voltage control problem as a CMDP. Section Ⅲ describes the framework of the proposed AL-SAC algorithm. Section Ⅳ presents and discusses the experimental results. Finally, Section Ⅴ concludes the paper.

## II. PROBLEM FORMULATION

This section will first introduce the concept of CMDP and then model the reactive voltage control problem in distribution networks as a CMDP.

### A. Preliminaries of Constrained Markov Decision Process

The Markov decision process (MDP) is widely adopted to formulate the sequential decision process with its predefined transition probability to model how the current state-action pair influences the state in the next step. By allowing for the inclusion of constraints that model the concept of safety, the MDP is further generalized as a constrained Markov decision process (CMDP). Generally, the CMDP can be represented by a tuple, which consists of a state space $S$, an action space $A$, a reward function $R$, a cost function $R^c$, a transition probability $Pr$, and a discount factor $\gamma \in [0,1]$.

Under a CMDP setting, an agent learns the optimal policy by interacting with the environment at each discrete time step, $t = 1, 2, ..., T$. In each time step $t$, the agent first observes the current state of the environment $s_t \in S$, then selects an action $a_t \in A$ guided by its policy. The action adopted at time step $t$ lead to the next state according to the unknown transition probability function $Pr(s_{t+1} | s_t, a_t)$. After the transition, the agent then receives the reward $R(s_t, a_t, s_{t+1}) \subset \mathbb{R}$ and the cost $R^c(s_t, a_t, s_{t+1}) \subset \mathbb{R}$ from the environment. In a CMDP, the reward and cost are associated with each action and state pair experienced by the agent, and the safety is maintained only if the expected discounted cost is below a certain threshold.

Instead of focusing on the reward and cost associated with individual action-state pairs, the agent aims to find the control policy $\pi$ that maximizes the long-term return and safety, which are represented by the expected discounted return $J(\pi)$ and the expected discounted cost $J^c(\pi)$:

$$\max_\pi J(\pi) \tag{1}$$

s.t.

$$J^c(\pi) \leq \bar{J} \tag{2}$$

where $\pi$ represents a mapping from a state space $S$ to an action space $A$ for a stochastic or deterministic policy. The expected discounted return is defined as $J(\pi) = E_{\tau \sim \pi}[\sum_{t=0}^{T} \gamma^t R_t]$, where $\tau$ is a trajectory or sequence of states and actions, $\{s_0, a_0, s_1, a_1, ... s_{T-1}, a_{T-1}, s_t\}$. $R_t$ is the short name for $R(s_t, a_t, s_{t+1})$. The expected discounted cost under policy $\pi$ in trajectory $\tau$ is defined in a similar manner: $J^c(\pi) = E_{\tau \sim \pi}[\sum_{t=0}^{T} \gamma^t R_t^c]$, where $R_t^c$ is $R^c(s_t, a_t, s_{t+1})$ for short.

The state-action value function $Q^\pi(s,a)$ is defined as follows:

$$Q^\pi(s,a) = E_{\tau \sim \pi}[\sum_{t=0}^{T} \gamma^t R_t \mid s_0 = s, a_0 = a] \tag{3}$$

where $Q^\pi(s,a)$ denotes the expected discounted reward starting from state $s$, taking action $a$, and thereafter following policy $\pi$. The value function satisfies the Bellman equation:

$$Q^\pi(s_t, a_t) = E_{a_{t+1} \sim \pi, s_{t+1} \sim Pr}[R_t + \gamma Q^\pi(s_{t+1}, a_{t+1})] \tag{4}$$

## B. Formulating Volt/Var Control Problem as CMDP

In the Volt/Var control problem, the distribution network's dispatch control center can be regarded as an agent interacting with the power grid system. By issuing control commands to the inverter-based energy resources (IB-ERs) in the grid, it optimizes both voltage levels and network losses. Consequently, the reactive power voltage control problem in the distribution network can be formulated as a CMDP. The state space, action space, reward function, and cost function are defined as follows:

1) State Space: the state space is defined as $s = [P, Q, V]$, where $P, Q, V$ represent the active power, reactive power, and voltage magnitude at the nodes, respectively.

2) Action Space: The action space is denoted as $a = Q_G$, where $Q_G$ represents the reactive power outputs of all IB-ERs and static Var compensators (SVCs). For IB-ERs, the range of $Q_G$ is constrained by $|Q_G| \leq \sqrt{S_G^2 - \overline{P_G}^2}$, where $S_G$ is the apparent power limit of the IB-ERs, and $\overline{P_G}$ is the active power output upper bound. For SVCs, the range of $Q_G$ is $\underline{Q_G} \leq Q_G \leq \overline{Q_G}$, where $\underline{Q_G}$ and $\overline{Q_G}$ represent the lower and upper bounds of the SVC output, respectively.

3) Reward and cost function: According to the definition of reinforcement learning, the reward $R(s_t, a_t, s_{t+1})$ and cost are $R^c(s_t, a_t, s_{t+1})$ functions of the state $s$. The objective of the reactive power voltage control problem is to minimize network losses while maintaining acceptable voltage levels. Therefore, the reward function is defined as:

$$R_t = -\sum P'$$

The definition of cost function is:

$$R_t^c = -\sum[\max(V' - \overline{V}, 0) + \max(\underline{V} - V', 0)]$$

when $R_t^c = 0$, it indicates that all node voltages are within acceptable limits, and when $R_t^c > 0$, it indicates that some node voltages violate the voltage level constraints.

Through the reward and cost functions defined above, the distribution network dispatch control center, replaced by a safe reinforcement learning agent, interacts with the power grid system and learns the optimal control policy. This enables the minimization of network losses while ensuring that voltage levels remain within acceptable limits.

## III. AUGMENTED LAGRANGIAN-BASED SAFE REINFORCEMENT LEARNING APPPROACH

In this section, we develop an innovative constrained deep reinforcement learning algorithm based on theaugmented Lagrangian method, thus we named it as augmented Lagrangian soft actor-critic (AL-SAC). To solve the examined problem, the adopted reinforcement learning algorithm should be sample efficient and constraints satisfied.

*Sample efficiency*: The training of DRL algorithms generally requires the agent to interact with the environment to collect sufficient samples. However, collecting a tremendous amount of operation experiences for EV charging scheduling problem repeatedly in real world is not realistic. Since the off-policy DRL algorithm's learned control policy (target policy) and the policy that generates instantaneous control action (behavior policy) are different, it allows the reuse historical operational experiences. Compared to on-policy ones, off-policy DRL algorithms are much more sample-efficient and appropriate for the examined problem.

*Constraints compliance*: Traditional RL methods have not considered safe exploration. While giving a RL agent complete freedom is unacceptable since certain exploratory behaviors may cause physical damage. For example, in the EV charging control problem, over-discharging will lead to significant SOC violations in the EV causing battery damage or dissatisfaction with the driver's charging demand. Thus, it is crucial to develop a RL algorithm, which can always achieve near constraint satisfaction.

In the following subsections, we first introduce the state-of-the-art maximum-entropy-based off-policy RL algorithm, soft actor-critic (SAC) [29]. Then we present the proposed constrained RL algorithm and discuss the algorithm design.

### A. Soft Actor-Critic

Actor-critic algorithms such as PPO [30], A3C [31], and DDPG [32] have been widely adopted in the DRL application. However, the first two are featured with sample inefficient, as they require samples generated by the latest policy at each gradient step. While the DDPG often suffers from hyperparameter sensitivity. To tackle the abovementioned challenges, [29] introduces the maximum-entropy concept in [33] into the actor-critic framework and develops the soft actor-critic (SAC) algorithm, which outperforms aforementioned algorithms.

By combining the entropy into the value function, SAC achieves a better tradeoff on exploration and exploitation and therefore avoids the suboptimal. The entropy for a probabilistic policy at state $s_t$ is defined as $H(\pi(\cdot | s_t)) = -\sum_a \pi(a | s_t) \log \pi(a | s_t)$. Then the state-action value function of SAC is defined as :

$$Q^\pi(s_t,a_t) = E_{a_{t+1}\sim\pi, s_{t+1}\sim Pr}[R_t + \gamma(Q^\pi(s_{t+1},a_{t+1}) + \alpha H(\pi(\cdot|s_{t+1})))] \tag{7}$$

The policy function is defined as a probability distribution $\pi(\cdot|s_t)$ in a stochastic manner as:

$$\pi(\cdot|s_{t+1}) = \frac{e^{\frac{Q^\pi(s,\cdot)}{\alpha}}}{\sum_a e^{\frac{Q^\pi(s,a)}{\alpha}}} \tag{8}$$

### B. Augmented Lagrangian SAC

Even though the SAC has achieved excellent performance on a range of challenging control tasks, it can only solve the MDPs. A popularly adopted approach to tackle CMDPs with DRL algorithms is revising the reward function via adding penalties associated with infeasible control over constraint. However, simply adding the product of fixed penalty coefficient and constraint violation into the reward function will lead to an infeasible or too consecutive control policy. In addition, identifying the penalty coefficient requires trial and error, which is low-efficient and time-consuming. In this subsection, we propose AL-SAC by extending the SAC algorithm to satisfy the operational constraints in CMDPs.

The optimal EV charging/discharging scheduling problem can be formulated as follows:

$$\max_\pi \mathcal{J}(\pi) = \mathbb{E}_{\tau\sim\rho_\phi}[\sum_{t=0}^T \gamma^t R_t] \tag{9}$$

s.t.

$$\underline{a} \leq a \leq \overline{a} \tag{10}$$

$$\mathbb{E}_{(s_t,a_t)\sim\rho_\pi}[-\log(\pi_i(a_t|s_t))] \geq \mathcal{H} \tag{11}$$

$$\mathcal{J}^c(\pi) = \mathbb{E}_{\tau\sim\rho_\phi}[\sum_{t=0}^T \gamma^t R_t^c] \leq \overline{\mathcal{J}_c} \tag{12}$$

where the objective function is to maximize the negative of charging cost. The first line of constraints denotes the action bound, in which the $\underline{a}$ and $\overline{a}$ denote the lower and upper bound, respectively. The second line of constraints is the lower bound of entropy and the third line is the upper bound for the SOC deviation. The threshold value of entropy is a hyperparameter which is used to balance the exploration and exploitation. For continuous action space, a general entropy threshold value is set as the negative value of the action space's dimension [29].

Note that the action bound has already been included in the action space's definition. We adopt the augmented Lagrangian method to transfer the original constrained optimization problem into the unconstrained formula:

$$\max_{\pi}\min_{\alpha,\lambda}\mathcal{J}(\pi)+\alpha(-\mathcal{H}-\log(\pi_i(a_t|s_t)))+\lambda(\bar{\mathcal{J}}_c-\mathcal{J}^c(\pi))+\frac{\delta_\lambda}{2}(\bar{\mathcal{J}}_c-\mathcal{J}^c(\pi))^2 \quad (13)$$

where $\alpha, \lambda$ are multipliers for entropy constraint and cost constraint, respectively. The $\delta_\lambda$ denotes the updating step size of $\lambda$, acts as the penalty coefficient in the augmented Lagrangian equation here. The entropy threshold and the upper bound for the SOC deviations are determined for specific optimization problems. In traditional RL-based methodologies, the SOC deviation is penalized directly in the reward function. In other word, they treat the variables $\alpha$ and $\lambda$ as penalty coefficients. However, the values for the penalty coefficients are hard to ensure. They can be either too conservative or infeasible. To overcome such a dilemma, we adopt the augmented Lagrangian method to solve the optimization problem, which guarantees that both primal and dual variables reach their optimal values and therefore guarantee the constraint compliance.

## C. Algorithm Design

Since the proposed AL-SAC is an off-policy DRL algorithm, parameters of actor and critic networks can be updated in an iterative manner using historical transitions. The overall framework of the AL-SAC is summarized in Algorithm 1. In each iteration, critic networks for evaluating expected discounted reward and cost are firstly trained by using stochastic gradient descent. Then, the Lagrange multipliers of corresponding constraints are updated using gradient ascent and the actor network is updated in a policy gradient manner with the unconstrained Lagrangian function.

---

**Algorithm 1:** Augmented-Lagrangian SAC Algorithm

1: **Initialization:** Total episodes: $N$, Time steps per episode: $T$, Replay Buffer: $\mathcal{D}$, Parameters of actor and critic networks and corresponding target copies: $\phi$, $\varphi$.

2: **for** $n = 0$ **to** $N$:

3:     **for** $t = 0$ **to** $T$:

4:         Output action $a_t$ after receiving state $s_t$;

5:         Receiving $R_t, R_t^c, s_{t+1}$ from the environment;

6:         Set $s_t = s_{t+1}$;

7:         Store transition $\{s_t, a_t, R_t, R_t^c, s_{t+1}\}$ in $\mathcal{D}$;

#Update Parameter

8:         Sample mini-batch transitions from $\mathcal{D}$;

9:         Update critic networks $\phi, \varphi$ by Eq.(16) and Eq.(19);

10:        Update Lagrange multipliers $\alpha, \lambda$ by Eq.(23) and Eq.(24);

11:        Formulate Lagrangian function as Eq.(21) and update actor network by Eq.(22);

12:        Update target networks in a soft manner.

| 13:     **end** |
| 14: **end** |

*1) Value function design and training*

In order to quantify the policies, the state-action value functions $Q^{\pi}(s,a)$ is defined as Eq.(9). In the Lagrangian function, the $Q^{\pi}(s,a)$ represents the expected discounted reward and the corresponding entropy multiplier product after taking action $a$ under state $s$ with policy $\pi$. Note that we store the tuple $\{s_t, a_t, R_t, R_t^c, s_{t+1}\}$ in the experience replay buffer $\mathcal{D}$ in each timestep and use these data for training.

We use two sets of neural networks to approximate the action-value function $Q^{\phi}(s,a)$ in timestep $t$. Using the Bellman equation, we could approximate the current state-action value with the expectation of all possible next state and corresponding actions with $\pi$. That is:

$$Q^{\phi}(s_t, a_t) \approx \mathbb{E}_{s,a,r,s_{t+1}}[y_t] \tag{14}$$

$$y_t = R_t + \gamma * Q^{\hat{\phi}}(s_{t+1}, a_{t+1}) \tag{15}$$

where $\phi$ is the target network, which is updated using soft update: $\phi = \eta\phi + (1-\eta\phi)$. The $\eta$ denotes the updating rate.

Hence the training for $\phi$ is to minimize the mean square error (MSE):

$$\mathcal{L}(\phi) = \mathbb{E}_{s,a,r,s_{t+1}}[(Q^{\phi}(s_t, a_t) - y_t)^2] \tag{16}$$

The action-state value function for the cost is designed in a similar manner. We use two sets of networks to approximate the cost value function $Q_c^{\varphi}(s_t, a_t)$. Similarly, the current cost value could be approximated as:

$$Q_c^{\varphi}(s_t, a_t) \approx \mathbb{E}_{s,a,r,s_{t+1}}[y_t^c] \tag{17}$$

$$y_t^c = R_t^c + \gamma * Q_c^{\tilde{\varphi}}(s_{t+1}, a_{t+1}) \tag{18}$$

where $\varphi$ is the target network for the cost value function and we update parameters of $\varphi$ by minimizing loss:

$$\mathcal{L}_c(\varphi) = \mathbb{E}_{s,a,r,s_{t+1}}[(Q_c^{\varphi}(s_t, a_t) - y_t^c)^2] \tag{19}$$

The cost value function networks also adopt the same soft update manner as the reward function networks.

*2) Policy function design and training*

As discussed previously, SAC adopts a stochastic policy, in which the policy $\pi$ is defined as a probability distribution $\pi(\cdot | s_t)$. Since direct optimization of a distribution is hard to implement, the policies $\pi$ is reparametrized as:

$$a_\theta(s_t, \xi_t) = \tanh(\mu_\theta(s_t) + \sigma_\theta(s_t) \odot \xi_t), \xi_t \sim \mathcal{N}(0,1) \tag{20}$$

where $\mu_\theta$ and $\sigma_\theta$ is the mean and standard deviation approximated by neural networks.

The goal of the policy network is to maximize the value of Lagrangian function. To drive the primal variable $\theta$ and dual variables $\alpha$ and $\lambda$ to reach the saddle-point $(\theta^*, \alpha^*, \lambda^*)$, we adopt the primal-dual method to alternatively update these variables.

With the defined value function, the original Lagrangian function is reformulated as follows:

$$L(\theta, \alpha, \lambda) = \mathbb{E}_{(s,a,r,s_{t+k}) \sim D}[Q^\phi(s_t, a_t)] + \alpha[-\mathcal{H} - \mathbb{E}_{(s_t,a_t) \sim D}[\log(\pi(a_t | s_t))]] \\ + \lambda\left[\overline{J_c} - \mathbb{E}_{(s,a,r,s_{t+k}) \sim D}[Q_c^\phi(s_t, a_t)]\right] + \frac{\delta_1}{2}[\overline{J_c} - \mathbb{E}_{(s,a,r,s_{t+k}) \sim D}[Q_c^\varphi(s_t, a_t)]]^2 \tag{21}$$

We update the primal variable $\theta$ at $i$-th iteration by using gradient ascent:

$$\theta_{i+1} = \theta_i + \delta_\theta \nabla_\theta L(\theta, \alpha, \lambda) \tag{22}$$

For the dual variables $\alpha$ and $\lambda$, we update them with the gradient ascent:

$$\alpha_{i+1} = [\alpha_i + \delta_\alpha \nabla_\alpha L(\theta, \alpha, \lambda)]_+ \tag{23}$$
$$\lambda_{i+1} = [\lambda_i + \delta_\lambda \nabla_\lambda L(\theta, \alpha, \lambda)]_+ \tag{24}$$

where the $[]_+$ is the projection to non-negative real numbers.

Note that the implementation details, such as the delayed update of the value function and the double-critics networks are omitted here.

## IV. CASE STUDIES

In this section, we apply the proposed AL-SAC algorithm for reactive voltage control across distribution networks of varying scales under real load fluctuation conditions. The performance of the proposed algorithm is evaluated by comparing it with methods proposed in previous studies.

### A. Environment Settings

The algorithm was tested on 33-bus, 69-bus, and 118-bus distribution networks to demonstrate the advantages of the proposed TC-DRL method. In the 33-bus test network, three IB-ERs with a reactive power capacity of 2 MVar and active power of 1.5 MW were connected at nodes 17, 21, and 24, while one SVC with a reactive power capacity of 2 MVar was connected at node 32. In the 69-bus test network, four IB-ERs with a reactive power capacity of 2 MVar and active power of 1.5 MW were connected at nodes 5, 22, 44, and 63, with one SVC of 2 MVar reactive power capacity connected at node 13. In the 118-bus test network, eight IB-ERs with a reactive power capacity of 2 MVar and active power of 1.5 MW were connected at nodes 33, 50, 53, 68, 74, 97, 107, and 111, with two SVCs of 2 MVar reactive power capacity connected at nodes 44 and 104. All load and generation data were consistent with the grid data in , provided by real-world grid measurement data with one data point every 15 minutes, yielding 96 data samples per day. The voltage limits for all nodes were set to [0.95, 1.05]. The test network data were downloaded from Matpower and imported into Pandapower . The algorithm was implemented in Python, using Pandapower to solve the balanced power flow for simulating the distribution network and PyTorch to implement the DRL algorithm.

*B. Benchmark Methods*

We selected several state-of-the-art deep reinforcement learning algorithms and model-based optimization methods as baseline approaches for comparison. These deep reinforcement learning algorithms include DDPG and SAC. All of these algorithms are capable of handling high-dimensional continuous state and action spaces. However, DDPG and SAC cannot be directly applied in a CMDP setting. To address the constraints in the problem under study, we manually selected penalty coefficients in the reward functions of these algorithms. It is important to note that the value of the penalty coefficients is difficult to determine directly and typically requires trial and error for evaluation. In our study, the reward functions of DDPG and SAC were defined as $R_t - \delta R_t^c$, in which the penalty coefficient $\delta$ is set as 5 and 0.5. Since DDPG, SAC, and AL-SAC are all off-policy deep reinforcement learning algorithms, the sample outcomes from each training iteration can be stored as historical data for model updates.

To better illustrate the effectiveness of the proposed AL-SAC algorithm, the neural network architecture and parameters of all these off-policy actor-critic DRL algorithms are set the same. The details of neural networks are presented in Table I. As discussed in Section III, to balance the exploration and exploitation, the value of target entropy is set as the negative value of action space's dimension.

TABLE I

Algorithm Hyperparameters

| Algorithm | Parameter | Value |
|---|---|---|
| DDPG, SAC and AL-SAC | Hidden layer size | [256,256] |
|  | Batch size | 256 |
|  | Discount factor $\gamma$ | 0.995 |
|  | Learning rate for network | 5e-4 |
| AL-SAC | Learning rate for multipliers $\alpha, \lambda$ | 1e-5 |
|  | Initial value for $\alpha, \lambda$ | 0 |
|  | Target Entropy $\mathcal{H}$ | -1 |

To demonstrate the convergence and effectiveness of all these deep reinforcement learning algorithms, we also employed a model-based optimization (MBO) algorithm. MBO utilizes an interior-point solver and an accurate power flow model to solve the reactive voltage control task. Its results can be considered as the optimal solution, serving as a benchmark for evaluating the performance of the DRL algorithms.

*C. Numerical Results of Centralized Framework*

Under the centralized control scheme, all reactive support devices (IB-ERs and SVCs) in the distribution network are scheduled and controlled by a reinforcement learning agent representing the distribution network control center. To verify the convergence and satisfaction of safety constraints of the proposed safe reinforcement learning method in the reactive voltage control task, this section analyzes the numerical results of various methods on the 33-bus and 69-bus systems. For the aforementioned reinforcement learning methods, the models are trained using 200 days of grid data. Since the dataset contains 96 data points per day, the training dataset consists of a total of 19,200 samples. To eliminate the impact of randomness caused by model training, all reinforcement learning algorithms are trained under three fixed random seed settings, and the average of the results is taken as the final model output.

Fig. 1-4 illustrates the training process and convergence of all reinforcement learning algorithms and the model-based optimization method on the 33-bus distribution network. Specifically, Figures 1 and 2 show the network losses caused by reactive voltage control across all methods on the 200-day training dataset, while Fig.3 and Fig.4 depict the voltage limit violations corresponding to the control strategies in the training dataset. For the reinforcement learning methods, the models tend to converge as the training steps increase. As shown in Fig. 1 and Fig. 2, the proposed AL-SAC algorithm converges in the latter half of the training process, with its control strategy resulting in network losses close to the theoretical optimal solution obtained by the model-based optimization method. In contrast, for the baseline reinforcement learning methods such as DDPG and SAC, it can be observed that these methods, which heavily rely on the selection of penalty coefficients in the reward function, struggle to effectively minimize network losses while simultaneously ensuring voltage limit compliance. Fig. 1 shows that when larger penalty coefficients are selected, the network losses caused by DDPG and SAC are higher than those of the AL-SAC algorithm, indicating that the strategies are overly conservative, prioritizing voltage limit compliance. Fig. 3 shows the voltage limit violations under various methods with larger penalty coefficients. It can be observed that the voltage violations caused by DDPG and SAC are relatively low, approximating those of the AL-SAC algorithm and the theoretical optimal result.

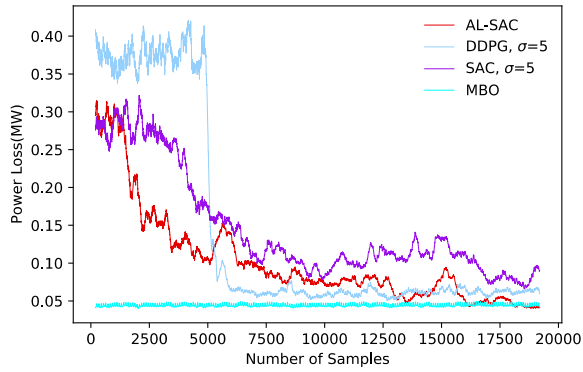

Fig. 1. Power loss in training process(33-bus)

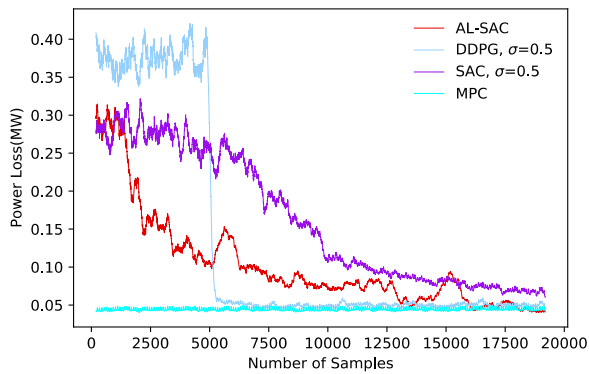

Fig. 2. Power loss in training process(33-bus)

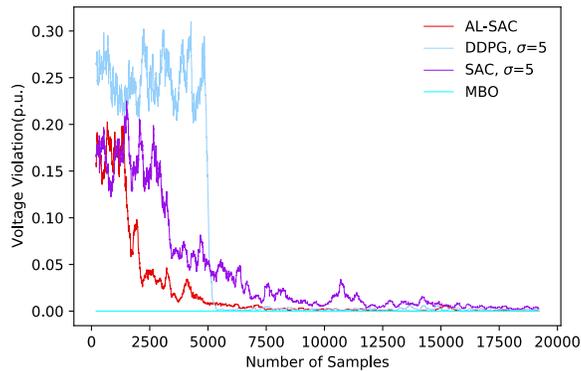

Fig. 3. Voltage Violation in training process(33-bus)

Fig. 2 demonstrates that when the penalty coefficients are reduced, the network losses caused by the DDPG and SAC methods decrease significantly, indicating that their current strategies prioritize minimizing network losses. As a deterministic strategy, DDPG even achieves lower network losses than the AL-SAC algorithm under these conditions. Correspondingly, in Fig. 4, while reducing the penalty coefficients leads to lower network losses for DDPG and SAC, it also results in severe voltage limit violations. Clearly, the proposed AL-SAC algorithm is capable of automatically deriving a reactive power control strategy that minimizes network losses while satisfying voltage limit constraints. In the absence of accurate distribution network model parameters, the solutions obtained by AL-SAC closely approximate the theoretical optimal results.

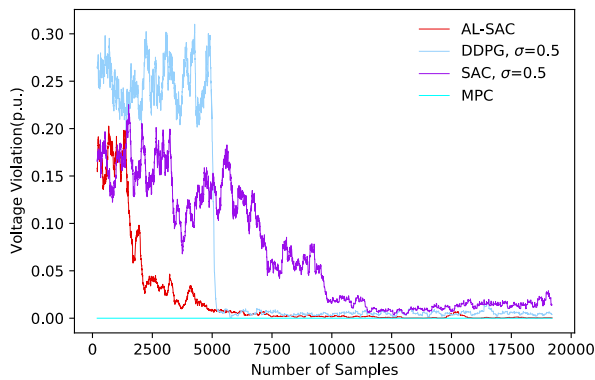

Fig. 4. Voltage Violation in training process(33-bus)

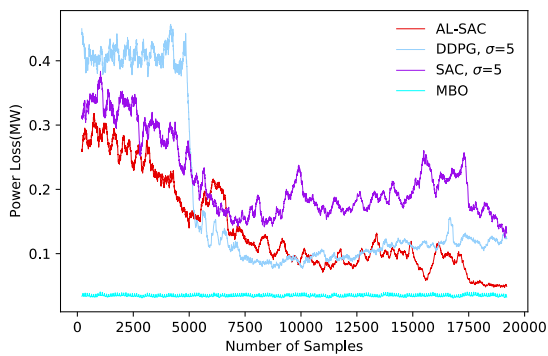

Fig. 5. Power loss in training process(69-bus)

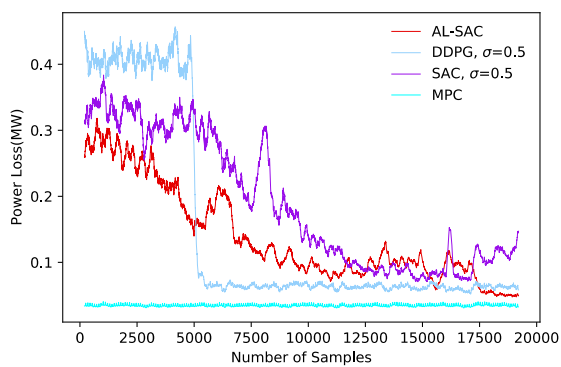

Fig. 6. Power loss in training process(69-bus)

Fig. 5-8 illustrates the training process and convergence behavior on the 69-bus distribution network. Similar to the results observed in the 33-bus system, reducing the penalty coefficients can lead to lower network losses, but at the cost of significant voltage limit violations. An interesting observation in the 69-bus system is that, even with larger penalty coefficients, the DDPG and SAC methods still result in severe

voltage limit violations. In contrast, the proposed AL-SAC algorithm effectively prevents voltage violations while achieving results that closely approximate the theoretically optimal solution.

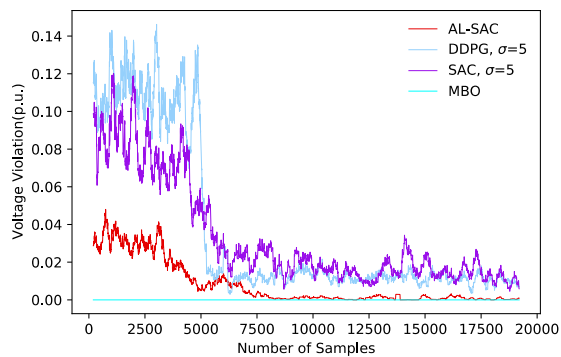

Fig. 7. Voltage violation in training process(69-bus)

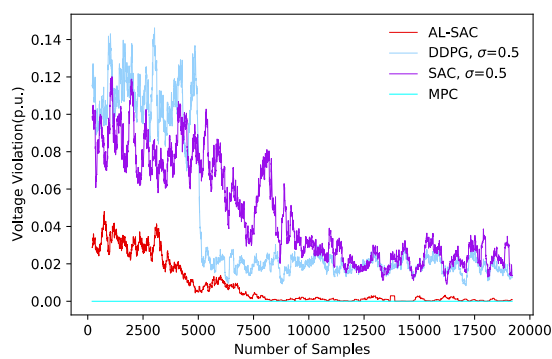

Fig. 8. Voltage violation in training process(69-bus)

TABLE II

ALGORITHM PERFORMANCE COMPARISON IN TEST DATASET

| Test Distribution System | Methods | Power Loss (MW) | Voltage Violation (p.u.) |
|---|---|---|---|
| 33-bus System | DDPG(penalty=0.5) | 0.0492 | 0.0041 |
| | DDPG(penalty=5) | 0.0631 | 0.0019 |
| | SAC(penalty=0.5) | 0.0661 | 0.0174 |
| | SAC(penalty=5) | 0.0802 | 0.0033 |
| | AL-SAC | 0.0426 | 0.0003 |
| | MBO | 0.0439 | 0 |
| 69-bus System | DDPG(penalty=0.5) | 0.0621 | 0.0157 |
| | DDPG(penalty=5) | 0.1216 | 0.0093 |
| | SAC(penalty=0.5) | 0.1480 | 0.0226 |
| | SAC(penalty=5) | 0.1219 | 0.0098 |
| | AL-SAC | 0.0501 | 0.0004 |
| | MBO | 0.0343 | 0 |

To provide a more intuitive representation of the performance of various methods in the reactive voltage control task for distribution networks, Table II quantitatively presents the network losses and voltage limit violations for AL-SAC, DDPG, SAC, and the model-based optimization method on the 33-bus and 69-bus systems. To validate the effectiveness of each approach, the load and generation data used for testing were not included in the training dataset. The test dataset consists of 10 days of data, with 96 data points per day. The trained AL-SAC, DDPG, and SAC models, obtained from the training dataset, were used to evaluate performance on the test dataset.

*D. Numerical Results of Decentralized Framework*

To demonstrate the scalability of the proposed reactive voltage control scheme, this paper further introduces a distributed reactive voltage control approach under a multi-agent setting. In larger-scale distribution networks, the system measurement information from each local area is delayed when uploaded to the control center. Distributed control agents can only observe the system measurements from adjacent subregions and then control the nearby reactive support devices (IB-ERs and SVCs) based on local observations. Under the distributed control framework, all agents collaborate to minimize network transmission losses while avoiding voltage limit violations.

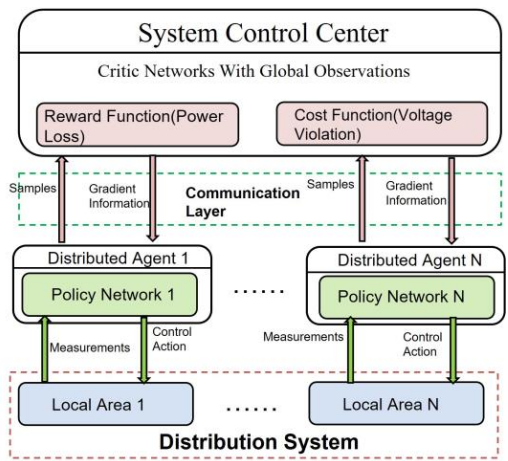

Fig. 9. Framework of decentralized Var/Voltage Control

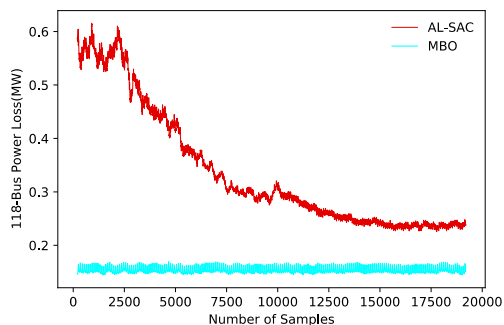

Fig. 10. Power loss in training process(118-bus)

To implement the distributed reactive voltage control, a centralized training and distributed execution strategy is adopted based on the proposed AL-SAC algorithm. The policy network for reactive power regulation in each local control area determines the corresponding reactive power adjustment actions upon receiving the local area measurement information and implements these actions. The local measurement information is then uploaded to the control center. The neural network for evaluating the reward function and cost function is trained on the global measurement dataset provided by the control center. Fig. 9 illustrates the framework of the extended multi-agent reactive voltage control scheme presented in this paper.

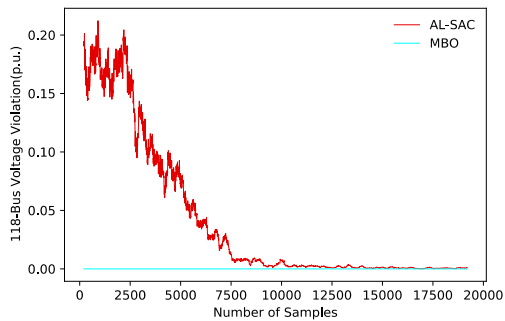

Fig. 11. Voltage violation in training process(118-bus)

To validate the proposed distributed reactive voltage control scheme under the multi-agent setting, experiments were conducted on the 118-bus distribution network system. Similar to the experimental setup for the 33-bus and 69-bus systems, the load and generation data were sourced from real grid measurements, spanning 200 days with 96 sample data points per day, resulting in a training set of 19,200 samples. In the 118-bus distribution network, each reactive compensation device is controlled by an AL-SAC agent, totaling 10 agents. Each distributed AL-SAC agent only receives the measurement information (including voltage magnitude, active power, and reactive power) from its respective node. Figure 10 shows the network losses of the AL-SAC algorithm compared to the theoretical optimal solution obtained by the model-based optimization method in the 118-bus system. Fig. 11 displays the corresponding voltage limit violations. The experimental results indicate that the proposed AL-SAC-based reactive voltage control algorithm demonstrates good scalability. Under the centralized training and distributed execution multi-agent framework, it achieves network losses close to the theoretical optimal solution while satisfying voltage level constraints. Table III presents the network losses and voltage limit violations of the trained distributed AL-SAC model on the same test dataset used in the previous experiments. The numerical results in Table III demonstrate that the proposed distributed reactive voltage control method for distribution networks achieves near-optimal control strategies, meeting the operational control requirements of the distribution network.

TABLE III
DECENTRALIZED ALGORITHM PERFORMANCE COMPARISON IN TEST DATASET

| Test Distribution System | Methods | Power Loss (MW) | Voltage Violation (p.u.) |
|---|---|---|---|
| 118-bus System | AL-SAC | 0.2374 | 0.0007 |

|  | MBO | 0.1524 | 0 |

## V. Conclusion

In this paper, we propose a model-free offline policy-based safe reinforcement learning algorithm, AL-SAC, to address the reactive voltage control problem in distribution networks. The algorithm is designed to minimize network losses while ensuring voltage level constraints are met, even in the absence of accurate distribution network model information. To prevent voltage violations, the reactive voltage control problem is formulated as a constrained Markov decision process. By incorporating the augmented Lagrangian method, the original constrained optimization problem is transformed into an unconstrained one. Through alternating updates of the neural network parameters and Lagrangian multipliers within the reinforcement learning model, the proposed safe reinforcement learning algorithm converges to the optimal solution. To validate the effectiveness and novelty of the proposed method, experiments were conducted on the 33-bus and 69-bus distribution networks. The experimental results demonstrate that the proposed method significantly outperforms other reinforcement learning approaches in reactive power optimization for distribution networks. To further verify the scalability of the proposed method, we introduced a centralized learning, distributed execution multi-agent framework, which successfully implemented distributed reactive voltage control in the 118-bus distribution network system.